\documentclass[runningheads]{llncs}
\usepackage{graphicx}
\usepackage{subcaption} 
\usepackage[T1]{fontenc}
\usepackage[utf8]{inputenc}
\usepackage{amsmath,amssymb,mathtools}
\usepackage{graphicx}
\usepackage{booktabs}
\usepackage{multirow}
\usepackage{makecell}
\usepackage{xcolor}
\definecolor{DeepGreen}{RGB}{40,160,4}
\definecolor{DeepRed}{RGB}{196,50,50}
\definecolor{DeepBlue}{RGB}{59,116,174}

\usepackage{colortbl}
\usepackage{microtype}
\usepackage{inconsolata}
\usepackage{float}
\usepackage{placeins}
\usepackage{subcaption}
\usepackage{wrapfig}   

\usepackage[T1]{fontenc}
\usepackage{graphicx}
\begin{document}
%
\title{Facet-Level Persona Control by Trait-Activated Routing with Contrastive SAE for Role-Playing LLMs}
\titlerunning{Facet-level Persona Control with Contrastive SAE for Role-Playing LLMs}
%
\author{First Author\inst{1}\orcidID{0000-1111-2222-3333} \and
Second Author\inst{2,3}\orcidID{1111-2222-3333-4444} \and
Third Author\inst{3}\orcidID{2222--3333-4444-5555}}
\author{
Wenqiu Tang\inst{1}\orcidID{0009-0008-6687-4460} \and
Zhen Wan\inst{2}\orcidID{0009-0008-4550-4794} \and
Takahiro Komamizu\inst{1}\orcidID{0000-0002-3041-4330} \and
Ichiro Ide\inst{1}\orcidID{0000-0003-3942-9296}
}

%
\institute{Nagoya University, Nagoya, Aichi, Japan \\
\email{tangw@cs.is.i.nagoya-u.ac.jp}, \email{taka-coma@acm.org}, \email{ide@i.nagoya-u.ac.jp}\\
\and
Kyoto University, Kyoto, Japan\\
\email{zhenwan.nlp@gmail.com}}

\maketitle              
\begin{abstract}

Personality control in Role-Playing Agents (RPAs) is commonly achieved via training-free methods that inject persona descriptions and memory through prompts or retrieval-augmented generation, or via supervised fine-tuning (SFT) on persona-specific corpora. While SFT can be effective, it requires persona-labeled data and retraining for new roles, limiting flexibility. In contrast, prompt- and RAG-based signals are easy to apply but can be diluted in long dialogues, leading to drifting and sometimes inconsistent persona behavior. To address this, we propose a contrastive Sparse AutoEncoder (SAE) framework that learns facet-level personality control vectors aligned with the Big Five 30-facet model. A new 15,000-sample leakage-controlled corpus is constructed to provide balanced supervision for each facet. The learned vectors are integrated into the model’s residual space and dynamically selected by a trait-activated routing module, enabling precise and interpretable personality steering. Experiments on Large Language Models (LLMs) show that the proposed method maintains stable character fidelity and output quality across contextualized settings, outperforming Contrastive Activation Addition (CAA) and prompt-only baselines. The combined SAE+Prompt configuration achieves the best overall performance, confirming that contrastively trained latent vectors can enhance persona control while preserving dialogue coherence.

\end{abstract}
\section{Introduction}
Personality in Role-Playing Agents (RPAs) —Large Language Models (LLMs) conditioned to enact consistent psychological traits of certain characters— has emerged as a primary driver of in-character coherence and user trust. Contemporary training-free RPA stacks typically combine carefully engineered personality system prompts and persona descriptors, Retrieval-Augmented Generation (RAG) for persona-relevant knowledge, while training-based RPA involves supervised finetuning on role-related corpus to internalize stylistic or behavioral constraints, enabling coherent role enactment beyond ordinary LLM chatbots~\cite{chen2024personapersonalizationsurveyroleplaying,wang-etal-2024-incharacter}. 

However, both RPA methods have limitations. Training-free pipelines can suffer from instruction-following degradation when long, role-related context (e.g., RAG passages) is injected, often introducing prompt noise and diluting the intended persona signal. Training-based RPA, in turn, typically demands substantial data and computation, and risks catastrophic forgetting of the model’s original general-purpose capabilities.
Such issues reflect broader challenges of maintaining controllable, interpretable generation of LLMs.
These limitations (Prompt brittleness/leakage long-context dilution, weak controllability, poor interpretability, and the cost/forgetting of training ) motivate a lightweight, interpretable inference-time control resilient to long context.

Many studies have shown that using Control Vectors (CVs) can effectively shape style and personality-like traits in LLMs' generations by adding small, attribute-aligned shifts to the residual stream, steering internal activations toward the desired direction~\cite{chen2025personavectorsmonitoringcontrolling,durmus2024steering,konen-etal-2024-style,turner2024-activation-addition} .
Although CVs have shown strong effects, several studies indicate instability in acquisition and usage; CVs derived from small or biased prompt-sets may capture prompt‐specific artifacts rather than generalizable semantic directions, yielding noisy or brittle interventions~\cite{braun2025understandingunreliabilitysteeringvectors,turner2024-activation-addition}.

To stabilize CVs for RPAs, we adopt a data-centric inference-time controller that couples facet-pure data with representation learning based on a Sparse AutoEncoder (SAE). First, a leakage-controlled Big Five personality~\cite{TettBurnett2003} corpus at the facet level supplies balanced positive/negative pairs and minimizes prompt/template artifacts. Next, an SAE yields a disentangled latent basis in which a contrastive objective learns compact within-facet and separated across-facet directions; these facet-aligned directions are injected as lightweight activation edits composed with standard RPA prompting without weight updates, improving stability, controllability, and interpretability under long context. 

Accordingly, our goal is to learn high-purity personality CVs effectively incorporated into existing RPA pipelines. Leveraging these CVs, we improve personality-simulation accuracy without sacrificing dialogue stability or role fidelity.
The contributions of this paper are as follows:
\begin{enumerate}
    
  \item \textbf{Dataset Construction} (Section~\ref{sec:dataset}): 
    We release a high-purity Big Five \emph{30-facet} corpus (15k instances, 500 per facet) designed to isolate single sub-dimensions, use first-person phrasing, and minimize cross-trait leakage.
  
    \item \textbf{Novel Framework} (Section~\ref{sec:prop}):
        We introduce a Contrastive-Learning (CL) objective for CVs in the SAE latent space, which promotes within-facet compactness and across-facet separation for more stable control. In addition, we employ an Agent-Based Decision Module to insert only the question-relevant CVs, thereby reducing noise. Together, these components achieve more discriminative and reliable personality steering.
    
   \item \textbf{Superior Performance} (Section~\ref{sec:eval}):
    We introduce contextual-based benchmark persona fidelity and conduct detailed analysis on five steered methods, where our CV-SAE+Prompt method attains the best overall results, indicating a robust way to steer RPA's personality.
\end{enumerate}

\section{Related Work}

Personality control methods can be roughly grouped into two families: \emph{Inference-time control}, and \emph{Training-time control}~\cite{chen2024personapersonalizationsurveyroleplaying,tseng-etal-2024-two}. 
We focus on \emph{Inference-time control} because it offers a strong trade-off between effectiveness and practicality. Unlike training-time methods, it does not require collecting large persona corpora or updating model weights, avoiding data costs, computational cost, and potential forgetting~\cite{Ding2023PEFT,keskar2019ctrlconditionaltransformerlanguage}. Within inference-time methods, prior work has explored decoding-time control, such as Plug-and-Play Language (PPLM)~\cite{Dathathri2020PPLM}), while effective, these methods incur substantial decoding-time compute, are sensitive to hyperparameters, and can sometimes degrade fluency or semantic fidelity~\cite{Dathathri2020PPLM}.In the contrast, Inference-time control using Control Vectors (CVs) or prompt methods~\cite{panickssery2024-caa,turner2024-activation-addition}, makes them lightweight, composable with standard decoders, and easy to ablate without retraining or modifying pretrained parameters.

There are two ways for generating CVs in the inference-time control: CV-SAE and CV-CAA. CV-SAE trains an SAE to obtain a disentangled feature space, then compute CVs from contrasts between persona-positive and negative samples~\cite{wang2025improvingllmreasoninginterpretable,Zhao2025SAE}. In the contrast, CV-CAA (CAA: Contrastive Activation Addition) directly computes a mean activation difference between paired prompts expressing opposite attributes~\cite{allbert2025identifyingmanipulatingpersonalitytraits,panickssery2024-caa}. Though simple and parameter-free, CVs' directions may overlap across traits, motivating orthogonalization or denoising to prevent cross-trait interference.

Unlike prior SAE/CAA pipelines, the proposed method (i) replaces generic templates with a professional, leakage-controlled 30-facet corpus, (ii) uses CL in SAE space to align and separate facet directions, and (iii) employs an agent policy that selects and schedules vectors from situational cues, rather than injecting all Big Five facet vectors at once.

\section{Big Five Dataset Construction}
\label{sec:dataset}
\begin{figure*}[t]
    \centering
    \includegraphics[width=1.1\textwidth]{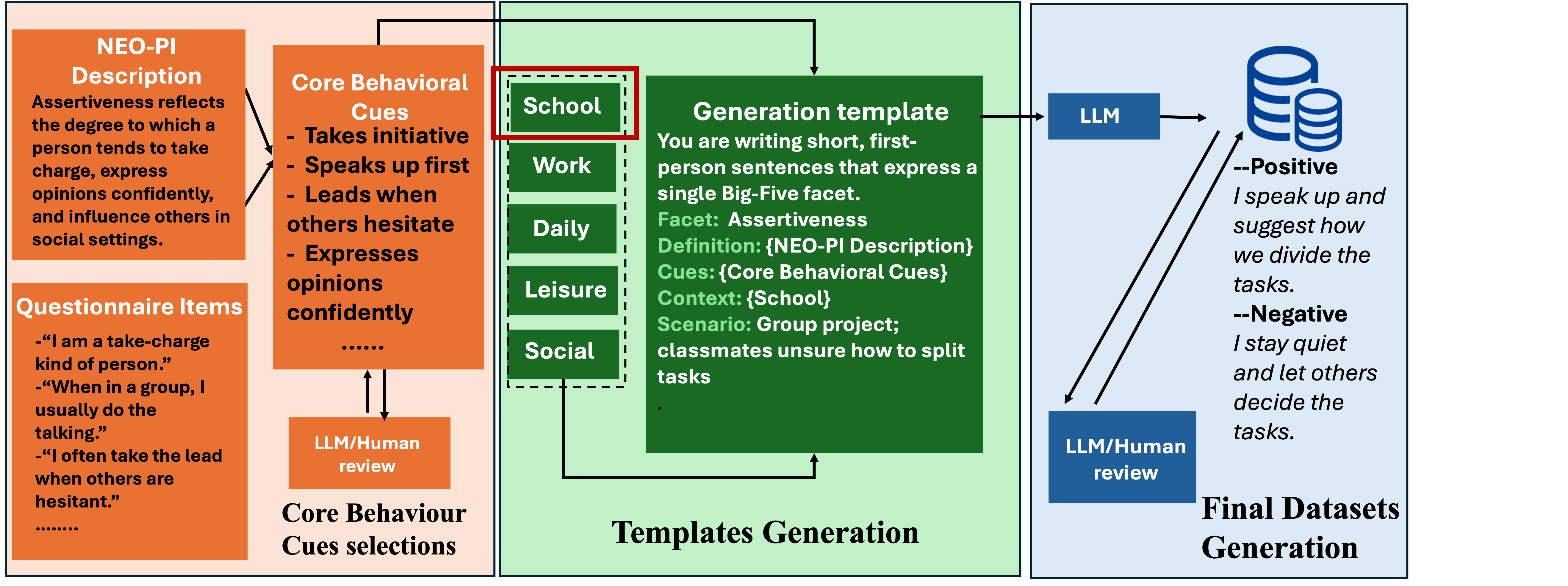}
    \caption{Pipeline for constructing the Big Five personality dataset, illustrated with the Assertiveness facet as an example.}
    \label{fig:BIG-5}
\end{figure*}

The Big Five personality traits form a psychological model that describes five broad dimensions of human personality.~\cite{TupesChristal1992}. We constructed a facet-level Big Five dataset based on the NEO Personality Inventory (NEO-PI)~\cite{costa2008revised} in a three-step pipeline shown in Fig.~\ref{fig:BIG-5}
NEO-PI splits each dimension in Big Five into six facets (30 in total). 
The constructed dataset consists of scenario-based positive and negative samples for each facet.

In the first step of the pipeline, for each facet, we summarize the official definition in NEO-PI and representative questionnaire items, and extract a small set of \textit{core behavioral cues} that represent the essential trait tendency for both positive and negative poles of each trait.
We then map these cues to everyday contexts (study, work, daily life, and social/leisure) and design brief \textit{scenario templates} (e.g., group work, receiving feedback, and helping a friend) that naturally evoke the target facet. This cue-to-scenario design grounds each instance in a realistic situation rather than an abstract self-report.
In the last step, using these templates, a GPT-5-class model~\cite{openai2025gpt5_1} generates short first-person sentences (maximum 18 words with simple vocabulary). For each scenario, the model produces a \textit{positive} sample expressing the facet and a \textit{minimal-edit negative} counterpart that keeps the same context but reverses the behavior. This yields 250 positive and 250 negative instances per facet (15,000 in total).

Finally, we adopt a three-step validation pipline. In the construction phase, (i) a secondary LLM screens \emph{all} items to check facet correctness, polarity clarity, and linguistic simplicity, and flags potentially problematic cases; (ii) we then randomly sample 50 items from each facet-specific subset and ask ten psychology graduate students (five female and five male in their twenties) to judge the same criteria, and compare their judgments with those of the secondary LLM on the sampled items to verify and calibrate the initial automatic screening. Items that are flagged in either step (i) or (ii) are revised or regenerated until no obvious facet or polarity errors remain. After this, (iii) we train a 30-way facet classifier on the resulting corpus and evaluate it on a held-out split, obtaining 78.4\% macro-F1 with only 6.2\% of misclassifications crossing Big Five dimensions, which indicates that samples are generally aligned with their intended facet and that cross-dimension leakage is limited. The final corpus is therefore \textit{substantially facet-consistent and context-grounded}, and well suited for studying fine-grained personality control and evaluation

\section{Contrastive Sparse AutoEncoder (SAE) Training and Agent-Based Decision Module}
\label{sec:prop}

\begin{figure*}[t]
    \centering
    \includegraphics[width=1.0\textwidth]{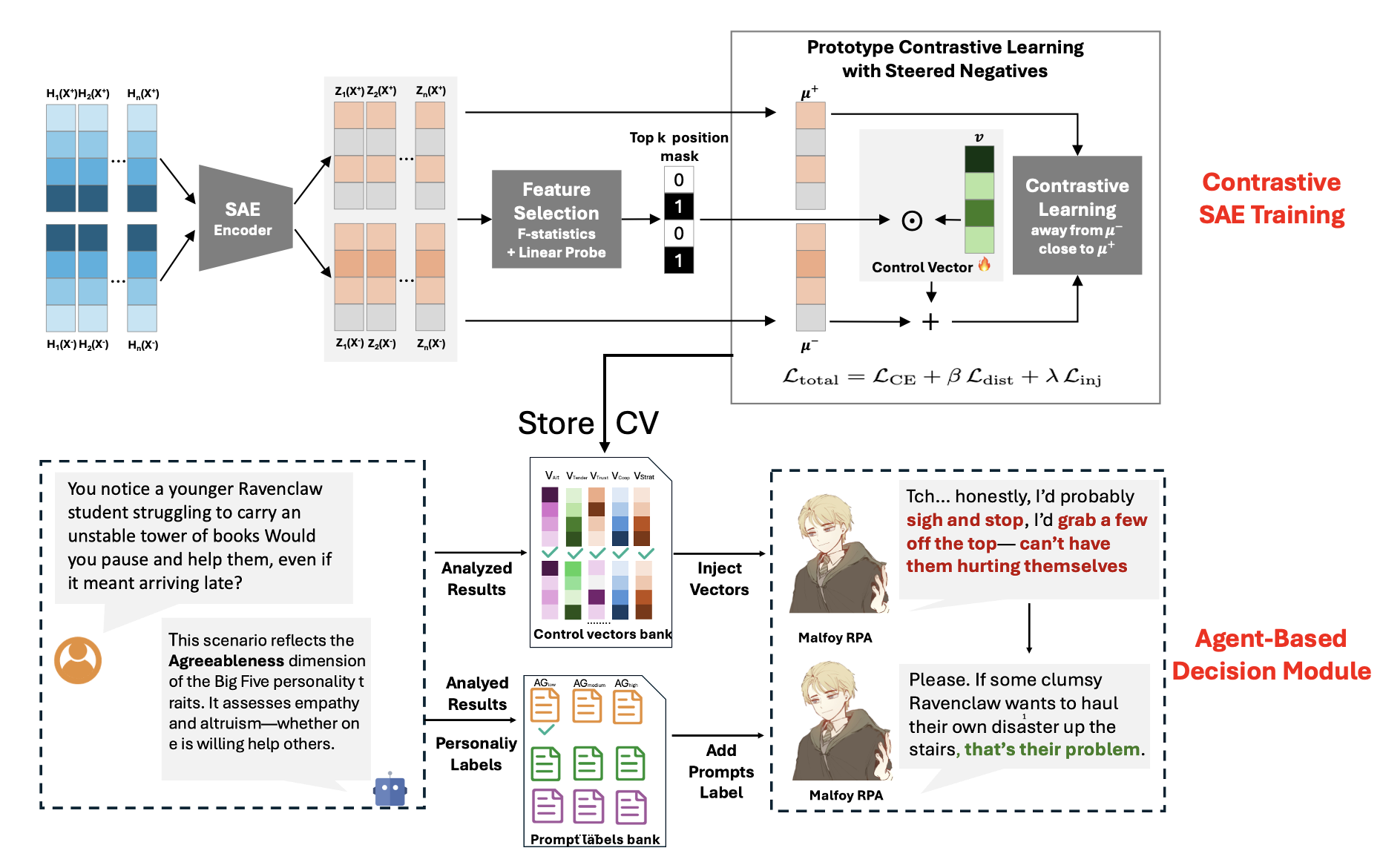}
    \caption{Proposed framework: Contrastive Sparse AutoEncoder (SAE) training and Agent-Based Decision Module for Control Vector (CV) injection.}
    \label{fig:sae-method}
\end{figure*}

Fig.~\ref{fig:sae-method} shows an overview of the proposed framework for steering an RPA’s personality expression by injecting learned CVs into the model’s hidden layers. It contains two steps: (i) SAE-based CV training mechanism enhanced with a Contrastive Learning (CL) loss to distill corpus-mined factors into facet-aligned CVs, which enhances effectiveness and strength in real-world applications, improving the model’s output quality along the intended control direction, and (ii) Self-reflection decision mechanism that dynamically selects which set of CVs to inject to precisely steer the models behaviours. 


\subsection{Contrastive SAE Training}



We map a hidden representation $\mathbf{h}(x)$ of $x$ into a pretrained SAE space to obtain a sparse latent code $\mathbf{z}$ through an encoding function $f_{\text{enc}}$ 
where each latent dimension is activated only by its associated concept in relevant contexts, thus yielding stronger control signals.
After training the encoder and decoder of an SAE on hidden states collected from the WikiText-103-raw-v1 corpus, 
a standard language-modeling dataset adopted in recent SAE studies, we decode the learned control vector $\mathbf{z}$ back into the model’s residual space as 
$\mathbf{v} = f_{\text{dec}}(\mathbf{z})$, and an injection is performed as:
\begin{equation}
\mathbf{h}'(x) = \mathbf{h}(x)+\alpha \mathbf{v}, 
\label{eq:injection}
\end{equation}
where 
\( \mathbf{v} \) is the CV, and 
\( \alpha \) controls the injection strength.

\label{subsubsec:contrastive}

Following prior work~\cite{he2025saessvsupervisedsteeringsparse}, we use F-statistics and linear probes to select the $d_{\text{steer}}$ ``top activated'' features ($d_{\text{steer}} \leq d$), where $d$ is the original dimensionality of the SAE space. After deciding these features, we construct a CV $\mathbf{v}\in\mathbb{R}^{d}$ that is nonzero only on the $d_{\text{steer}}$ selected coordinates. We initialize it in the SAE space using the class-centroid difference as:

\begin{equation}
\mathbf{v} = (\boldsymbol{\mu}^{+} - \boldsymbol{\mu}^{-}) \odot \mathbf{m},
\label{eq:CVs}
\end{equation}
where $\boldsymbol{\mu}^{+}$ and $\boldsymbol{\mu}^{-}$ are the mean SAE codes for positive and negative samples from the generated BFI dataset, and $\mathbf{m} \in \{0, 1\}^d$ masks the active coordinates.

The total objective to learn the CVs is a linear combination of 
a prototype contrast loss $\mathcal{L}_{\text{CE}}$,
a positive pull loss $\mathcal{L}_{\text{dist}}$, and
a regularization factor as:
\begin{equation}
\label{eq:total}
\mathcal{L}_{\text{total}}
= \mathcal{L}_{\text{CE}}
+ \beta\,\mathcal{L}_{\text{dist}}
+ \lambda \lVert \mathbf{v} \odot \mathbf{m} \rVert_2^2 ,
\end{equation}
where $\beta$ and $\lambda$ are tunable weight parameters.

\textbf{Prototype contrast loss $\mathcal{L}_{\text{CE}}$}: 
Given a CV \textbf{v}, our goal is to \emph{pull representations toward the positive centroid and push them away from the negative one}. CL naturally enforces this push--pull geometry~\cite{pmlr-v119-chen20j,HU2024128645}, yielding discriminative facet directions.
For a negative sample with SAE code $\mathbf{z}$, we inject a masked vector $\mathbf{v}$ and $\ell_2$-normalize it as:
\begin{equation}
\mathbf{z}^{+} = \frac{\mathbf{z} + \mathbf{v}}{\lVert \mathbf{z} + \mathbf{v} \rVert_2}.
\label{eq:normalize}
\end{equation}
We likewise normalize 
$\tilde{\boldsymbol{\mu}}^{+} = \boldsymbol{\mu}^{+}/\lVert \boldsymbol{\mu}^{+}\rVert_2$ and
$\tilde{\boldsymbol{\mu}}^{-} = \boldsymbol{\mu}^{-}/\lVert \boldsymbol{\mu}^{-}\rVert_2$
for stability.
Based on them, cosine scores are defined as
$c_{+}=\langle \mathbf{z}^{+},\tilde{\boldsymbol{\mu}}^{+}\rangle$ and
$c_{-}=\langle \mathbf{z}^{+},\tilde{\boldsymbol{\mu}}^{-}\rangle$.
Following ArcFace/CosFace~\cite{Deng_2022,wang2018cosfacelargemargincosine}, we impose an angular margin on the positive side and
a cosine margin on the negative side as:
\begin{equation}
\label{eq:margins}
\tilde c_{+}=\cos\!\big(\arccos(c_{+}) + m_{\text{pos}}\big),
\quad
\tilde c_{-}=c_{-}+m_{\text{neg}},
\end{equation}
where $m_{\text{pos}}>0$ and $m_{\text{neg}}>0$ are margins.
Using scale $s>0$ (equivalently, temperature $\tau=1/s$),
we form logits $\ell=[s\tilde c_{+},\, s\tilde c_{-}]$ and minimize the
two-way cross-entropy loss that treats the positive prototype as the target as:
\begin{equation}
\label{eq:ce}
\mathcal{L}_{\text{CE}}
= -\frac{1}{B}\sum_{i=1}^{B}
\log\frac{\exp(s\tilde c_{+,i})}
{\exp(s\tilde c_{+,i})+\exp(s\tilde c_{-,i})}.
\end{equation}
\textbf{Positive pull loss $\mathcal{L}_{\text{dist}}$}: 
We use a distance-based loss that compares the injected representation to the class centroids in the active subspace as:
\begin{equation}
\label{eq:dist}
\mathcal{L}_{\text{dist}}(v)
= \big\| \mathbf{z}^{+} - \mathbf{u}^{+} \big\|_{2}
- \big\| \mathbf{z}^{+} - \mathbf{u}^{-} \big\|_{2},
\end{equation}
where $\mathbf{z}^{+}$ is the post-injection representation, and $\mathbf{u}^{+}$ and $\mathbf{u}^{-}$ are the positive and negative centroids, respectively. 


\subsection{Agent-Based Decision Module}

Guided by Trait Activation~\cite{TettBurnett2003}, we activate \emph{only} the trait cued by the current prompt (e.g., writing advice \(\rightarrow\) \emph{Openness}, not \emph{Extraversion}), avoiding competition between CVs. Accordingly, we select best\mbox{-}matching CVs per turn; persona knowledge come from RAG/prompt conditioning, while the vector acts as an \emph{amplifier}, routing by contextual cues.
To realize this idea, we use the role\mbox{-}play model’s base LLM as a routing agent. Before producing the final answer, as shown in Figure~\ref{fig:sae-method}, the agent receives the user’s query and infers which facet-level CVs are most strongly cued by the question.
Based on the agent’s analysis, we inject CVs corresponding to the most relevant facets for the given prompt in each dimension which is composed of six facet-level CVs. This process allows fine\mbox{-}grained steering while keeping inference\mbox{-}time control targeted.



\section{Experiments}\label{sec:eval}


In personality measurement, \emph{abstract (i.e., non-contextualized)} items are global self-descriptions posed without an explicit frame of reference, whereas \emph{contextualized} items incorporate a frame of reference by stipulating the setting/role/time~\cite{PaulhusJohn1998}.  We followed the \textit{In-Character} benchmark, which evaluates 44 abstract Big Five open-ended questions and uses GPT-4o~\cite{openai2024gpt4o} as the judge~\cite{wang-etal-2024-incharacter}. However, abstract Big Five questions can elicit socially desirable responses~\cite{PaulhusJohn1998}. 
Grounded in reference-group effects and trait-activation theory~\cite{TettBurnett2003}, 
we rewrote all 44 items into \emph{contextualized questions} using GPT-4o, adding minimal temporal/situational cues to 
reduce bias while preserving semantics.

\subsection{Evaluation Protocol}
We evaluated on two compact, open-weight instruction followers that are widely used in RPA pipelines (e.g., RoleLLM): 
\texttt{Qwen/Qwen3-4B-Instruct-2507} (Qweb3-4B for short)~\cite{yang2025qwen3technicalreport} 
and \texttt{mistralai/Mistral-7B-Instruct-v0.7} (Mistral-7B for short)~\cite{jiang2023mistral7b}. 
Qwen3-4B delivers strong small-scale performance, while Mistral-7B provides efficient attention and competitive results at the 7B scale.

In order to observe the behaviors of RPAs, 
this experiment reports four evaluation metrics: Full-Accuracy (FA), Mean Squared Error (MSE), Mean Absolute Error (MAE), and Muti-Turn Rate (MTR).
FA, MSE and MAE evaluate correctness in estimating characters: FA is the proportion of characters (out of 26) for which all five Big Five dimensions are predicted correctly, and MSE and MAE are per-dimension errors and stability across paraphrased questions. MTR evaluates whether the agent maintains the intended persona while answering, and whether the response contains any of the errors including repetition, out of characters, and multi-turn dialogue, which are also evaluated by LLM.

\subsection{Implementation and Comparison Methods}
\label{sec:pipelines}

In the proposed method (called CV-SAE), we injected CV at the mid-residual-layer with the strongest stylistic effects, specifically the 15th layer in Qwen3-4B~\cite{yang2025qwen3technicalreport} and the 16th layer in Mistral-7B~\cite{jiang2023mistral7b}, aligned with prior findings~\cite{sae_finetuning,panickssery2024-caa} and our preliminary results. 
SAE is trained at the same layer.

To validate our proposal, we have the following experiment setups.
\textit{Base RPA} denotes no additional intervention.
\textit{Prompt-Label} denotes conditional generation on explicit Big Five~\cite{TupesChristal1992} labels using per-dimension low/medium/high facet-covering descriptors chosen by the character’s overall score~\cite{chen2024usingpromptsguidelarge}.
\textit{CV-SAE} and \textit{CV-CAA} denote injecting CVs derived from an SAE and CAA, respectively. 
\textit{Combined Methods} (CV-CAA+Prompt and CV-SAE+Prompt) are RPAs that combine CV injection methods with Prompt-Label to assess their complementarity.

\begin{table}[t]
\centering
\caption{
Performance of Qwen3-4B~\cite{yang2025qwen3technicalreport} and Mistral-7B~\cite{jiang2023mistral7b} on Big Five personality questions. 
Best results for each metric are \textbf{bolded}. 
}
\resizebox{0.9\linewidth}{!}{%
\label{tab:exp-performance}
\centering
\small
\begin{tabular}{lrrrrrrrr}
\toprule
\multirow{2}{*}{Method}
& \multicolumn{4}{c}{Abstract Questions} 
& \multicolumn{4}{c}{Contextual Questions} \\
\cmidrule(lr){2-5} \cmidrule(lr){6-9}
&
FA{\tiny $(\uparrow)$} & 
MSE{\tiny $(\downarrow)$} & 
MAE{\tiny $(\downarrow)$} & 
MTR{\tiny $(\downarrow)$} &
FA{\tiny $(\uparrow)$} & 
MSE{\tiny $(\downarrow)$} & 
MAE{\tiny $(\downarrow)$} & 
MTR{\tiny $(\downarrow)$} \\
\midrule

\multicolumn{9}{c}{Qwen3-4B} \\
\midrule
\multicolumn{9}{l}{\bf Baselines} \\ 
Base RPA       & 7.7 & 16.0 & 31.3 & \textbf{0.0} & 7.7 & 14.2 & 30.2 & 0.1 \\
Prompt-Label   & 50.0 & 10.9 & 24.9 & \textbf{0.0} & 30.7 & 11.7 & 26.2 & 0.1 \\
CV-CAA         & 73.1 & 4.8 & 16.1 & \textbf{0.0} & 76.9 & 4.2 & 14.7 & \textbf{0.0} \\
\cmidrule(lr){1-9}
\multicolumn{9}{l}{\bf Proposed} \\ 
CV-SAE & 76.9 & 4.9 & 16.5 & \textbf{0.0} & 76.9 & 3.4 & 14.3 & \textbf{0.0} \\
\cmidrule(lr){1-9}
\multicolumn{9}{l}{\bf Combined} \\ 
CV-CAA+Prompt     & \textbf{81.0} & 3.7 & 14.2 & \textbf{0.0} & 81.0 & 3.4 & 14.1 & 0.2 \\
CV-SAE+Prompt     & \textbf{81.0} & \textbf{3.2} & \textbf{14.1} & \textbf{0.0} & \textbf{88.5} & \textbf{2.4} & \textbf{12.1} & 0.1 \\

\midrule

\multicolumn{9}{c}{Mistral-7B}  \\
\midrule
\multicolumn{9}{l}{\bf Baselines} \\ 
Base RPA       & 19.2 & 15.5 & 32.8 & 5.6 & 3.9 & 17.8 & 34.7 & 1.4 \\
Prompt-Label   & 42.3 & 9.5  & 24.8 & \textbf{1.6} & 42.3 & 9.5  & 24.8 & \textbf{1.3} \\
CV-CAA         & 76.9 & 4.2  & 15.6 & 29.1 & 84.7 & 4.2 & 16.6 & 25.6 \\
\cmidrule(lr){1-9}
\multicolumn{9}{l}{\bf Proposed} \\ 
CV-SAE & 73.1 & 5.4 & 18.7 & 7.0 & 80.8 & 3.97 & 15.1 & 3.1 \\
\cmidrule(lr){1-9}
\multicolumn{9}{l}{\bf Combined} \\ 
CV-CAA+Prompt     & 38.5 & 9.3 & 23.8 & 1.8 & 46.2 & 10.0 & 24.6 & 1.4 \\
CV-SAE+Prompt     & \textbf{84.6} & \textbf{3.8} & \textbf{14.9} & 4.2 & \textbf{88.5} & \textbf{3.1} & \textbf{13.7} & 3.8 \\

\bottomrule
\end{tabular}
}
\end{table}

\subsection{Results}
\label{sec:results}

Table \ref{tab:exp-performance} summarizes the experimental results.
First, both CV methods (CV-SAE, CV-CAA) substantially outperformed the Prompt-Label baseline across FA, MSE, and MAE, confirming the effectiveness of direct vector modulation over prompt-only conditioning.
However, CV-CAA showed unstable behavior on Mistral-7B~\cite{jiang2023mistral7b}, where its MTR increased sharply, indicating that contrastive activation cues could destabilize dialog flow.
In contrast, CV-SAE achieved a stronger overall balance ---Improving accuracy while maintaining response smoothness--- suggesting that SAE yielded more stable and disentangled CV injection.
In particular, integrating prompts with CVs amplifies the contrast between the two approaches.
The combined CV-SAE+Prompt consistently attained the highest FA and the lowest reconstruction errors (MAE/MSE) across both backbones, pushing the overall performance to a new peak.
Conversely, CV-CAA+Prompt severely collapsed on Mistral-7B (FA dropping from 76.9 to 38.5), showing that prompt signals interfere with contrastive activations rather than reinforcing them.
On Qwen3-4B, this interference was less pronounced, yet SAE-based combined method remained superior in both accuracy and stability.

CV methods were resilient —and often improved— under contextualized questions.
On Mistral-7B, FA increased for CV-SAE, CV-CAA, and CV-SAE+Prompt.
On Qwen3-4B, CV-SAE+Prompt gained +7.5 pp, while CV-SAE/CAA held steady or improved slightly.
By contrast, Prompt-Label alone did not generalize as well: on Qwen, FA dropped from 50.0 (abstract) to 30.7 (contextual), suggesting that explicit labels without latent steering struggled to adapt to situational cues.

\captionsetup[subfigure]{labelformat=simple,labelsep=space}
\renewcommand\thesubfigure{(\alph{subfigure})}

\begin{figure*}[t] 
  \centering

  \begin{subfigure}[t]{0.48\linewidth}
    \centering
    \includegraphics[width=\linewidth]{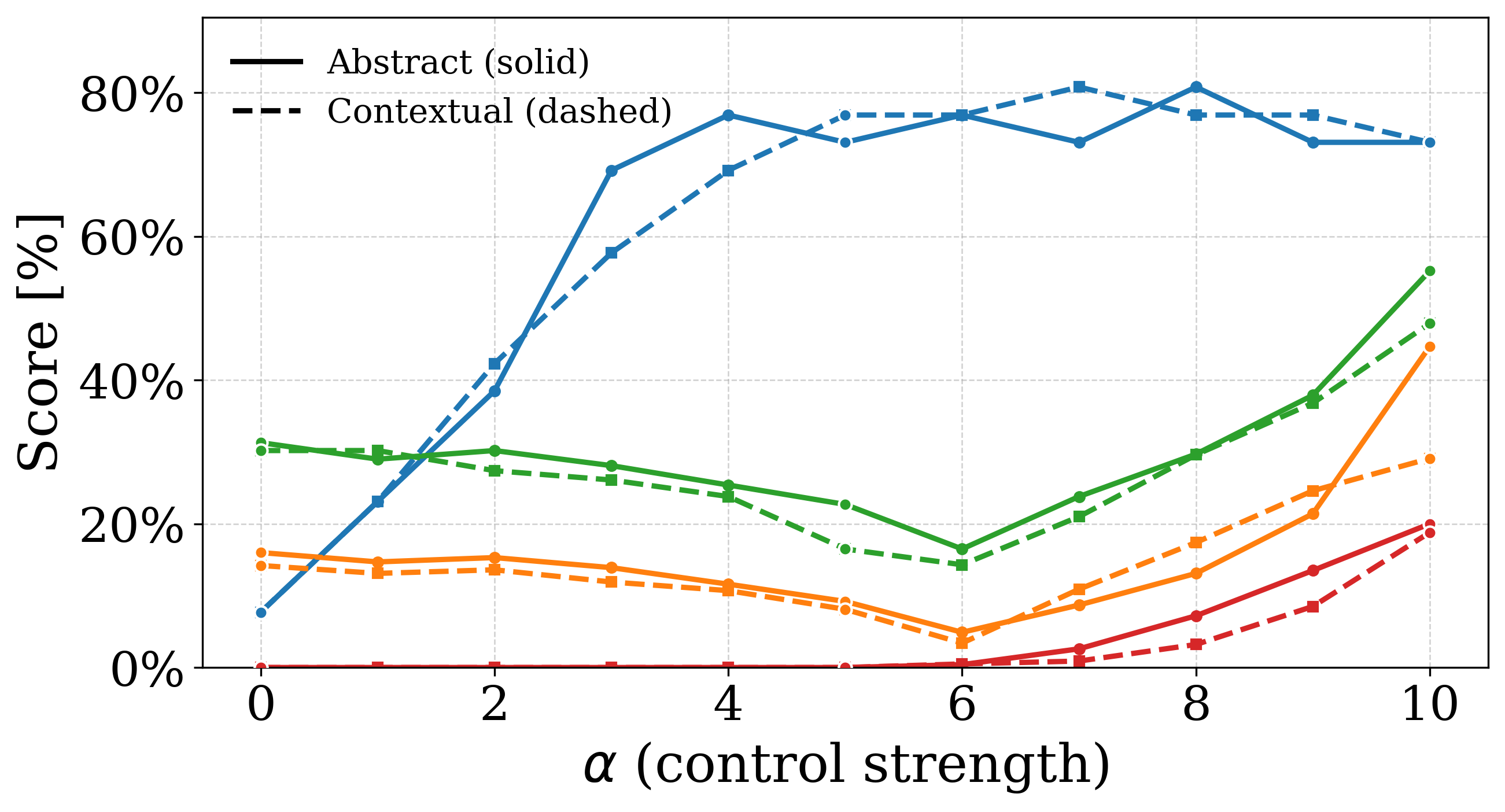}
    \vspace{-2em}
    \caption{CV-SAE with Qwen3-4B~\cite{yang2025qwen3technicalreport}}
    \label{fig:qwen-sae}
  \end{subfigure}\hfill
  \begin{subfigure}[t]{0.48\linewidth}
    \centering
    \includegraphics[width=\linewidth]{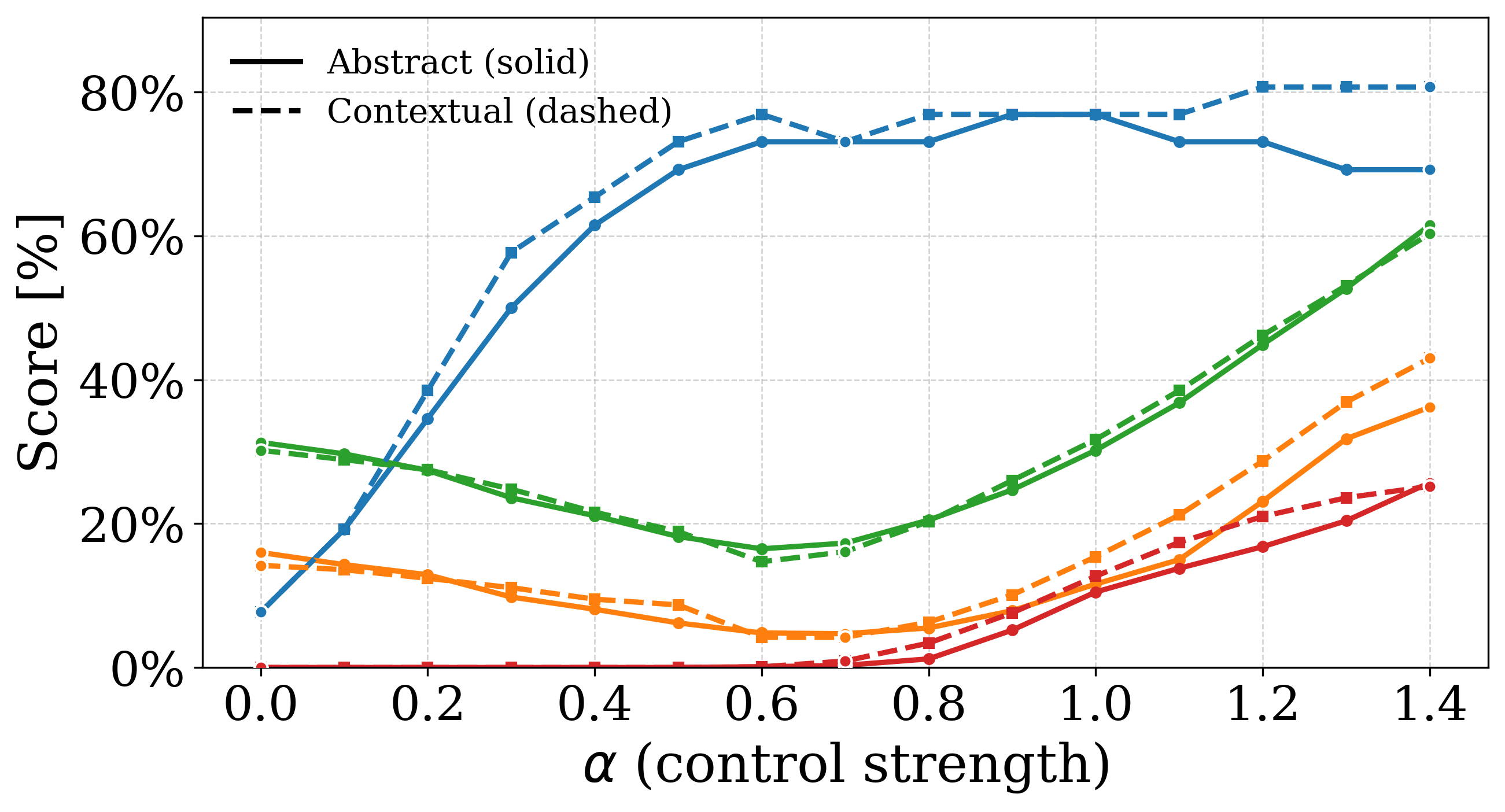}
    \vspace{-2em}
    \caption{CV-CAA with Qwen3-4B~\cite{yang2025qwen3technicalreport}}
    \label{fig:qwen-caa}
  \end{subfigure}

  \vspace{0.4em} 

  \begin{subfigure}[t]{0.48\linewidth}
    \centering
    \includegraphics[width=\linewidth]{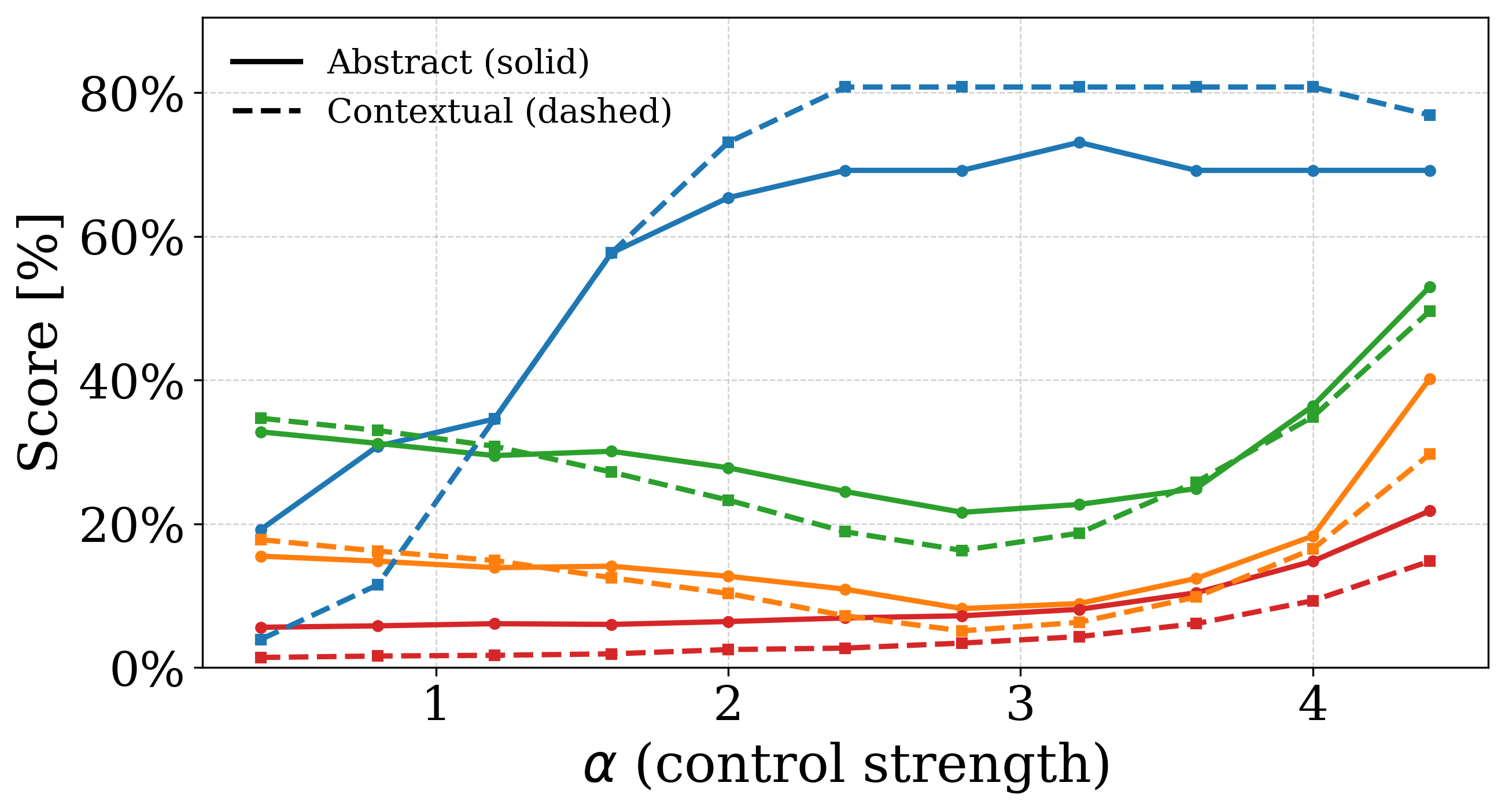}
    \vspace{-2em}
    \caption{CV-SAE with Mistral-7B~\cite{jiang2023mistral7b}}
    \label{fig:mistral-sae}
  \end{subfigure}\hfill
  \begin{subfigure}[t]{0.48\linewidth}
    \centering
    \includegraphics[width=\linewidth]
    {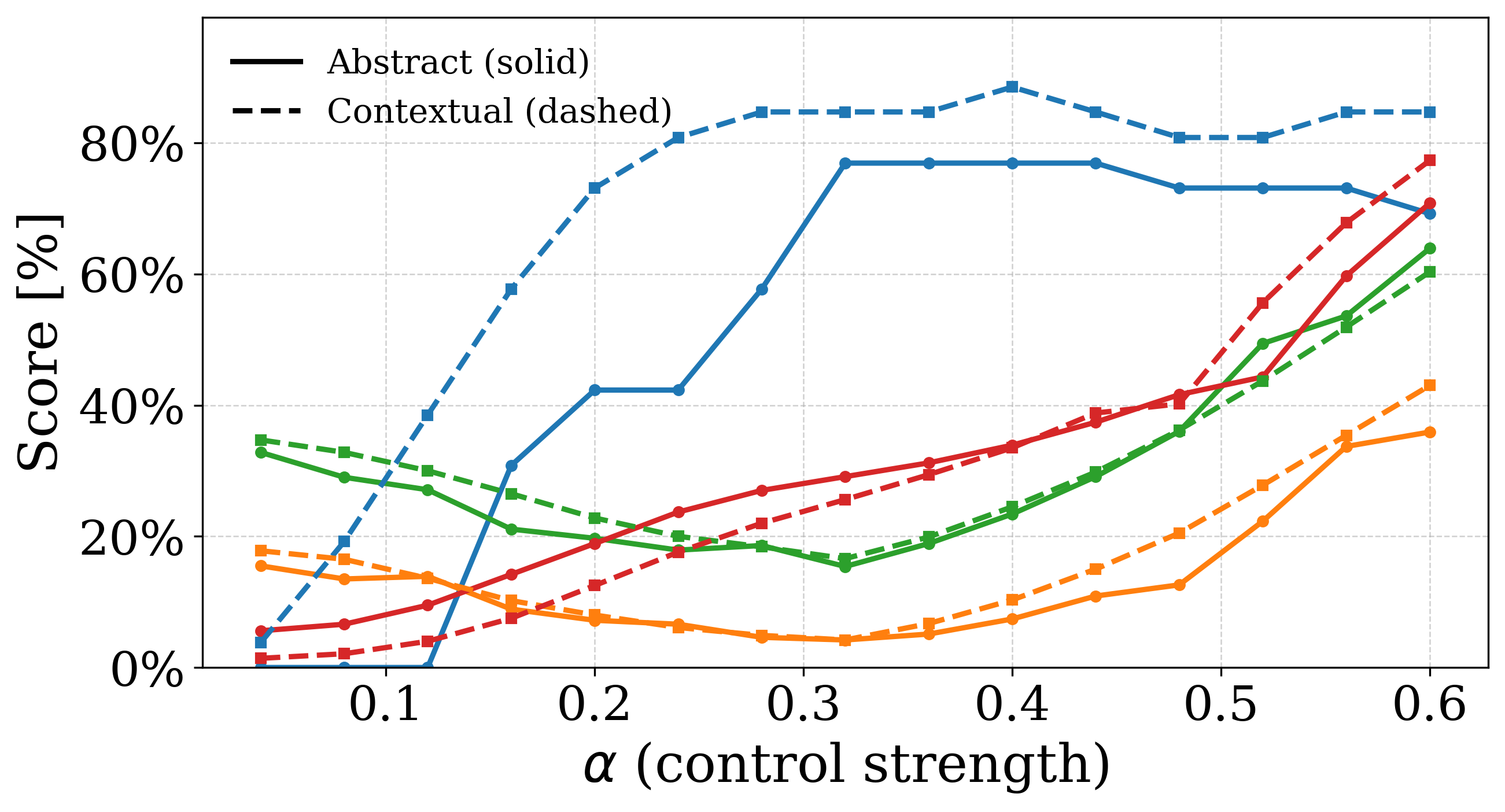}
    \vspace{-2em}
    \caption{CV-CAA with Mistral-7B~\cite{jiang2023mistral7b}}
    \label{fig:mistral-caa}
  \end{subfigure}

  \caption{
Effect of control strength $\alpha$ on Role-Play Agent (RPA) performance across backbones and control methods.
Results are reported under the \emph{abstract} (solid) and \emph{contextual} (dashed) settings.
\textcolor{DeepBlue}{Blue} denotes FA, 
\textcolor{orange}{orange} denotes MSE,
\textcolor{DeepGreen}{green} denotes MAE, and
\textcolor{DeepRed}{red} denotes MTR.
}

  \label{fig:overview-four-panels}
\end{figure*}

\subsection{Effect of Injection Strength $\alpha$}
As described in Eq.~\ref{eq:injection}, \(\alpha\) scales the CV. A too small value yields under-steering, while a too large value perturbs the semantic space and degrades responses. 
To identify a stable operating point, we swept \(\alpha\) and reported the performance. As shown in Fig.~\ref{fig:overview-four-panels}, mid-range \(\alpha\) values with CV-SAE delivered high FA and low errors with minimal MTR, whereas CV-CAA peaked at a smaller \(\alpha\) and showed faster MTR growth as \(\alpha\) increased. 

Increasing \(\alpha\) sharply inflated the three stability indicators, MSE, MAE, and MTR; In interview settings, the RPA’s personality became inconsistent even within the same question set (MAE and MSE rose faster), and output coherence as well as instruction following deteriorated (MTR kept increasing). 
Accordingly, we tuned \(\alpha\) on a preferring mid-range values for SAE and smaller values for CAA to balance accuracy and stability.

Across the four graphs in Fig.~\ref{fig:overview-four-panels}, CV-CAA exhibited a consistently higher (worse) MTR than SAE on both Qwen3-4B~\cite{yang2025qwen3technicalreport} and Mistral-7B~\cite{jiang2023mistral7b}, indicating more dialogue drift and less stable answers.
Since CV-SAE injects CVs only in the middle layers, whereas CV-CAA injects a small control signal into \emph{every} layer, the way of injection affects the performance.
We hypothesize that the control signal learned by CV-CAA was less ``pure''; As $\alpha$ increased, off-target semantic components were introduced alongside the intended concept direction, which degraded the response quality and inflated  MTR.
In contrast, CV-SAE provided a more concentrated ``purer'' steering direction; It carried less extraneous content and thus amplified the intended concept more reliably, leading to a higher FA without a rise in MTR.

\begin{table}[t]
\centering
\caption{
Effect of Contrastive Learning (CL) ---
The left block reports cosine similarities between the injected representation
\(z\) and the positive/negative Extraversion centroids \((\mu^+,\mu^-)\) on the
Extraversion subset of our corpus. The right block reports full-accuracy (FA)
on Abstract (Abst.) and Contextual (Cont.) Extraversion questions for ten
role agents under three training settings.
.
}
\resizebox{0.9\linewidth}{!}{%
\label{tab:effects}
\begin{tabular}{l rlrlrlrl rlrlrlrl}
\toprule
& \multicolumn{8}{c}{Similarities to centroids}
& \multicolumn{8}{c}{Full-accuracy}
\\ 
\cmidrule(lr){2-9}
\cmidrule(lr){10-17}

& \multicolumn{4}{c}{Qwen3-4B~\cite{yang2025qwen3technicalreport}} 
& \multicolumn{4}{c}{Mistral-7B~\cite{jiang2023mistral7b}}
& \multicolumn{4}{c}{Qwen3-4B~\cite{yang2025qwen3technicalreport}} 
& \multicolumn{4}{c}{Mistral-7B~\cite{jiang2023mistral7b}} \\
\cmidrule(lr){2-5}\cmidrule(lr){6-9}
\cmidrule(lr){10-13}\cmidrule(lr){14-17}
& \multicolumn{2}{c}{$\langle \mathbf{z}, \boldsymbol{\mu}^+ \rangle$}
& \multicolumn{2}{c}{$\langle \mathbf{z}, \boldsymbol{\mu}^- \rangle$}
& \multicolumn{2}{c}{$\langle \mathbf{z}, \boldsymbol{\mu}^+ \rangle$} 
& \multicolumn{2}{c}{$\langle \mathbf{z}, \boldsymbol{\mu}^- \rangle$} 
& \multicolumn{2}{c}{Abst.} & \multicolumn{2}{c}{Cont.}
& \multicolumn{2}{c}{Abst.} & \multicolumn{2}{c}{Cont.} \\
\midrule

Before training
& 0.92 & 
& 0.84 &
& 0.64 &
& 0.38 & 
& 11.5 & & 11.5 & 
& 23.1 & & 3.9 & \\

Train w/o CL   
& 0.76 & ($\downarrow$)  
& 0.31 & ($\downarrow$)  
& 0.29 & ($\downarrow$)  
& $-0.01$ & ($\downarrow$) 
&  0.0 & ($\downarrow$) &  3.9 &($\downarrow$)
&  3.9 & ($\downarrow$) &  0.0 & ($\downarrow$) \\

Train w/ \; CL      
& 0.93 & ($\uparrow$)
& 0.59 & ($\downarrow$) 
& 0.75 & ($\uparrow$) 
& 0.21 & ($\downarrow$)
& 76.9 & ($\uparrow$)   & 80.7 & ($\uparrow$)
& 76.9 & ($\uparrow$)   & 84.6 & ($\uparrow$) \\

\bottomrule
\end{tabular}
}
\end{table}

\subsection{Effect of CL} 
To further understand the effect of CL, we conducted an ablation study under three training settings---\textit{Before Training}, \textit{Training without Contrastive Learning}, and \textit{Training with CL} (see Table~\ref{tab:effects}). In each setting, we (i) computed the cosine similarity between the injected representation $z$ and the positive/negative Extraversion centroids $(\mu^+, \mu^-)$ estimated from the Extraversion subset of our corpus, and (ii) evaluated full-accuracy on ten representative role agents (selected from 26 personas) using both Abstract and Contextual items.
 
As shown in Table~\ref{tab:effects}, CV-SAE relying solely on the distance-based ($\ell_2$) loss indeed reduced the cosine similarity to the negative centroid $\boldsymbol{\mu}^{-}$ on both Qwen3-4B~\cite{yang2025qwen3technicalreport} and Mistral-7B~\cite{jiang2023mistral7b}. 
However, this also undesirably decreased the similarity to the positive centroid $\boldsymbol{\mu}^{+}$, respectively, deviating from the intended training objective. 
Correspondingly, Table~\ref{tab:effects} shows that CV-SAE without CL sharply dropped the overall full-accuracy, even lower than the untrained baseline. 
In contrast, introducing CL not only increased $\langle \mathbf{z}, \boldsymbol{\mu}^{+} \rangle$ while suppressing $\langle \mathbf{z}, \boldsymbol{\mu}^{-} \rangle$ but also lead to a substantial full-accuracy improvement across both abstract and contextual questions  confirming that contrastive optimization effectively aligned latent representations with the intended trait polarity.

\section{Conclusion}
We introduced an SAE-based contrastive CV for RPA personality steering. It uses a facet-accurate personality corpus to compute latent centroids and a reflection module to select relevant CVs. Combined with top-k latent filtering and contrastive tuning at mid-residual layers, the method achieves stable facet-level personality control. On Qwen3-4B~\cite{yang2025qwen3technicalreport} and Mistral-7B~\cite{jiang2023mistral7b}, SAE vectors improved accuracy and reduced error with minimal drift, outperforming CAA and prompting while preserving persona knowledge for interpretable long-context role-play.

\bibliographystyle{splncs04}
\bibliography{custom}





\end{document}